\documentclass[letterpaper, 10 pt, conference]{ieeeconf}  

\IEEEoverridecommandlockouts                              
                                                          
\usepackage{epsfig}         
\usepackage{amsmath}        
\usepackage{amssymb}        
\usepackage{graphicx}
\usepackage{pgfplots}
\pgfplotsset{compat=1.18} 
\usepackage{cite}
\usepackage{tabularx} 
\usepackage{booktabs}
\usepackage{url}
\usepackage{caption}
\title{\LARGE \bf
Stage-Transition Dense Reward Modeling for Reinforcement Learning
}

\author{
Yang Yang$^{1,\dagger}$, Bingjie Chen$^{1,\dagger}$, Zihan Wang$^{1}$, 
Yizhe Li$^{2}$, Guoping Pan$^{2}$, Yi Cheng$^{2}$, Houde Liu$^{1,2,*}$
\thanks{
$^{1}$Tsinghua Shenzhen International Graduate School,
Tsinghua University, Shenzhen 518055, China.
}
\thanks{
$^{2}$Zerith Robotics, Shenzhen 518055, China.
}
\thanks{
\makebox{$^{\dagger}$}Equal contribution.\protect\\
\makebox[1.5em][l]{}E-mails: Yang Yang (\texttt{yang-y25@mails.tsinghua.edu.cn}),\protect\\
\makebox[1.5em][l]{}Bingjie Chen (\texttt{bingjie.chenn@gmail.com}).
}
\thanks{
$^{*}$Corresponding author: Houde Liu. }
\thanks{
This work was supported by the Shenzhen Science and Technology Program 
(Grant No. RCJC20210706091946001) and the Shenzhen Science and Technology Program 
(Grant No. ZDCY20250901104207008).
}
}

\begin{document}
\maketitle
\thispagestyle{empty}
\pagestyle{empty}

\begin{abstract}
Reinforcement learning for long-horizon robotic manipulation is often limited by sparse and delayed rewards, while manually designing dense shaping signals is costly and brittle to changes in environments and object configurations. This work proposes Stage-Transition Dense Reward (STDR), a visual reward-learning framework that converts unstructured expert videos into logically grounded dense rewards for training RL agents from scratch. STDR leverages semantic understanding to infer a task’s stage structure from demonstrations, and delivers two complementary learning signals during online training: (i) stage-transition feedback that provides goal-directed reward, and (ii) within-stage progress feedback that supplies fine-grained guidance toward completing each stage. Furthermore, an out-of-distribution (OOD) detection mechanism and a grasping regulation module are integrated to enhance robustness and prevent reward hacking.  Experiments on 14 manipulation tasks across MetaWorld, ManiSkill, and Franka Kitchen show that STDR consistently improves sample efficiency and success rates over multiple baselines, and matches or surpasses handcrafted dense rewards on several challenging tasks. Real-robot evaluations further indicate that STDR assigns stable, progress-aligned rewards on successful executions while producing appropriately low rewards for failures, suggesting robustness to visual noise and better-calibrated reward assignment across settings.

\end{abstract}

\section{Introduction}
Reinforcement Learning (RL) has shown strong potential for robotic manipulation \cite{levine2016end, luo2024hilserl}. However, many practical manipulation problems are still challenging because reward feedback is often sparse, delayed, or only available upon task completion. This issue becomes especially severe in long-horizon settings \cite{nair2018overcoming}, where a successful episode requires completing a sequence of semantically different subtasks. With only a terminal success signal, learning frequently suffers from inefficient exploration and large interaction budgets \cite{riedmiller2018learning}.

A common remedy is to design dense rewards that provide intermediate guidance. Yet dense reward engineering is notoriously time-consuming, highly task-specific, and brittle to changes in environments or object configurations \cite{ng1999policy}. These limitations have motivated learning reward functions from expert demonstrations \cite{abbeel2004apprenticeship}, so that the agent can receive informative feedback throughout an episode. In principle, demonstration-based reward learning reduces manual design effort and offers a scalable path to tackle sparse-reward tasks. In practice, however, most readily available demonstrations---especially videos---are \emph{unstructured}: they do not come with explicit subtask boundaries or progress annotations. As a result, turning raw expert videos into a reliable, temporally coherent, and \emph{dense} learning signal remains a key bottleneck.

\begin{figure*}[t] 
    \centering
    \includegraphics[width=1\textwidth]{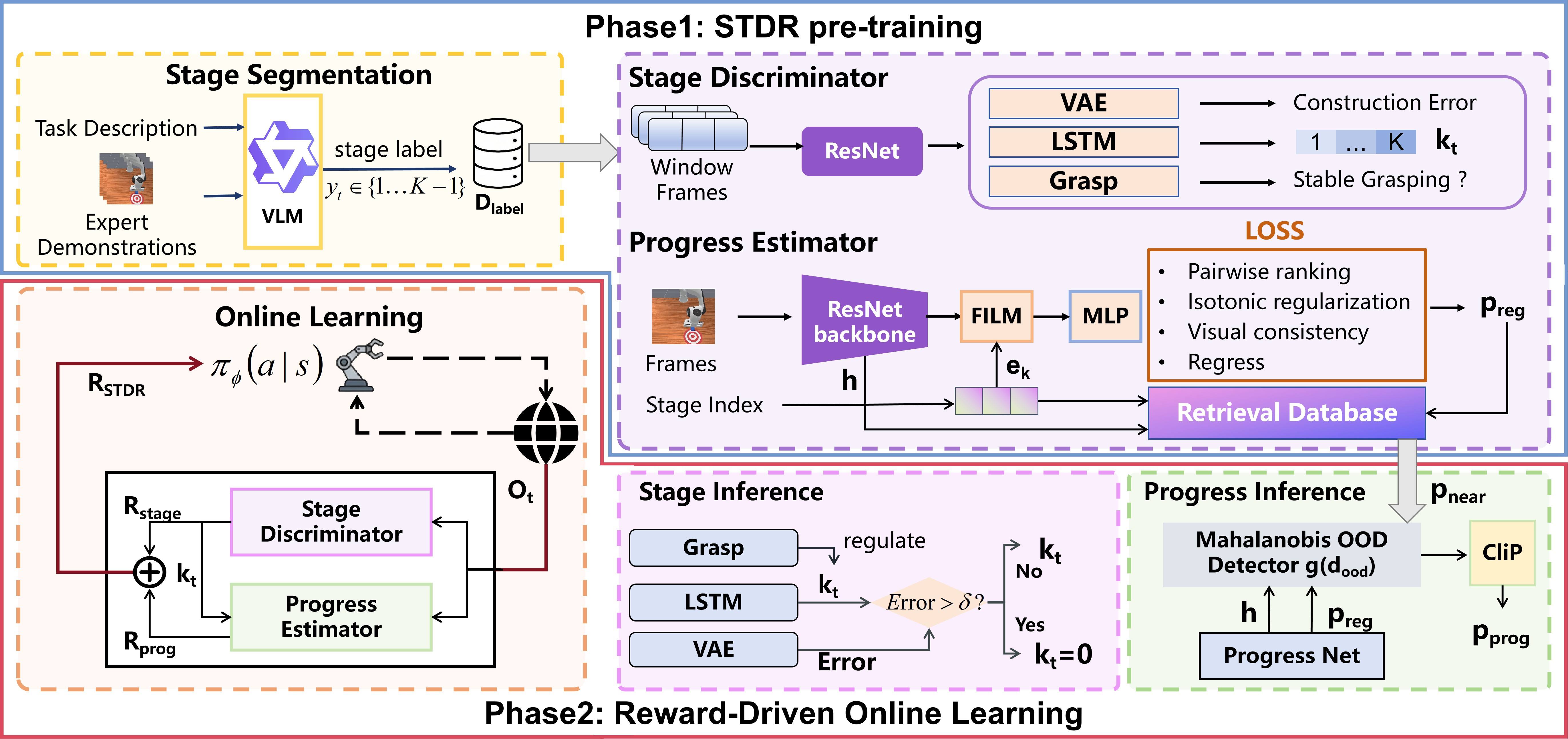}
    \caption{Overview of the STDR framework. (a) STDR pre-training: Expert videos are segmented by a VLM to generate pseudo-labels $y_t$ for establishing a labeled dataset $D_{label}$. These labels supervise the training of the Reward Model, which integrates a Stage Discriminator to output stage indices $k_t$ and a Progress Estimator conditioned on FiLM-based stage embeddings $e_k$. (b) Reward-Driven Online Learning: The pre-trained STDR module processes real-time continuous observation streams to generate a composite dense reward $R_{STDR}$, enabling efficient policy optimization via reinforcement learning.}
    \label{fig:framework}
\end{figure*}

A natural way to impose structure on long-horizon manipulation is to view it as progressing through multiple functional stages (e.g., approach, grasp, insert), where each stage corresponds to a meaningful sub-goal. If such stage structure and within-stage progress can be inferred from visual demonstrations, then the agent can be trained with rewards that reflect how far it has advanced, rather than only whether it has succeeded at the end. Building on this idea, we propose STDR, a framework that distills structured guidance from unstructured expert videos and produces dense reward signals for training RL agents from scratch, as shown in Fig.\ref{fig:framework}. At a high level, STDR leverages semantic understanding to obtain stage information from demonstrations, then provides two complementary forms of feedback during RL: a {coarse} signal that reflects stage transitions across the full task horizon, and a \emph{fine-grained} signal that measures progress within the current stage. To improve reliability beyond the expert distribution, STDR further incorporates an OOD aware mechanism so that the reward remains meaningful when observations drift away from the demonstration manifold.

Our main contributions are as follows:
\begin{itemize}
    \item We propose STDR, a vision-based reward learning framework that distills stage-structured dense rewards from unstructured expert videos via semantic task staging and online stage inference.
    \item We propose a structured reward shaping principle for long-horizon manipulation: combining global, stage-monotonic guidance (stage-transition feedback) with local, within-stage dense gradients (progress feedback), yielding an incentive landscape that is both temporally coherent and fine-grained.
    \item We introduce robust mechanisms against out-of-distribution states and reward hacking, including OOD-aware stage gating, OOD-gated progress fusion, and grasp-transition verification, and validate their gains across multiple tasks with ablations.
\end{itemize}

\section{Related Work}

\subsection{Reward Learning from Demonstrations}
Learning reward functions from expert behavior has been extensively studied in inverse reinforcement learning (IRL) \cite{liu2022inferring, chen2023fast}, which seeks to recover reward functions that explain demonstrations, reducing reliance on manually engineered shaping. However, in robotics, observations are often high-dimensional and partially observable, making reward inference and generalization challenging without strong inductive biases or access to privileged state features. Consequently, recent research has explored learning reward signals directly from expert visual observations, enabling reward modeling in end-to-end visual RL settings. Representative methods such as Value-Implicit Pre-training (VIP) \cite{ma2022vip} and Language-Image Representations and Rewards (LIV) \cite{ma2023liv} learn task-agnostic reward surrogates by embedding observations into latent spaces and defining goal-reaching objectives in that space. While effective for short-horizon skills, such similarity-based rewards can struggle in long-horizon manipulation tasks where progress depends on temporal task logic and compositional subtasks rather than purely geometric proximity. In contrast, STDR decomposes the reward signal into functional segments, ensuring a more monotonic and directed incentive structure.


\subsection{OOD Robustness and Verification}
Reward learning methods, including IRL-style approaches, are vulnerable under distribution shift: a policy may visit out-of-distribution (OOD) states and exploit spurious reward correlations, leading to reward hacking and unreliable optimization \cite{amodei2016concrete}. Prior work improves robustness by broadening supervision, such as incorporating sub-optimal or ranked trajectories as in T-REX \cite{brown2019extrapolating}, or by constraining early exploration through policy initialization, e.g., behavioral cloning \cite{rajeswaran2017learning, escoriza2025multistage}. Despite these efforts, reliable reward evaluation remains difficult when critical interaction cues occupy only a small portion of the image and the agent deviates from the expert distribution. Our method addresses this challenge by introducing an explicit verification mechanism to improve reward reliability under OOD observations.

\section{METHODOLOGY}

\subsection{Problem Formulation}
We model the robotic task as a Markov Decision Process (MDP) \cite{bellman1957markovian}, defined by the tuple $(\mathcal{S}, \mathcal{A}, P, R, \gamma)$. Here, $\mathcal{S}$ denotes the state space, $\mathcal{A}$ the action space, $P(s' \mid s, a)$ the transition dynamics, $R$ the reward function, and $\gamma \in [0,1)$ the discount factor. The goal of the RL agent is to find an optimal policy that maximizes the expected cumulative return, $
J(\pi)=\mathbb{E}_{\pi}\left[\sum_{t=0}^{\infty}\gamma^t R(s_t,a_t)\right]$. In most scenarios, the environment provides only a sparse reward signal, denoted as $R_{\text{sparse}}$, which is typically defined as a binary indicator that yields 1 only upon successful task completion and 0 otherwise. However, for many manipulation tasks, relying solely on such sparse feedback often fails to guide the agent toward the goal, or requires a prohibitive number of environment interactions before any meaningful reward signal can be observed. To address this challenge, our objective is to learn a dense reward function directly from expert demonstrations. The learned reward function is then used to train an RL agent from scratch, thereby facilitating more efficient policy optimization and improving sample efficiency in long-horizon tasks.

\subsection{Stage Segmentation}
To learn a structured reward from unstructured expert videos, we first conceptualize the long-horizon task as a sequence of semantically distinct stages. Drawing inspiration from recent advances in vision-language reasoning, we leverage a VLM-Qwen3~\cite{qwen3} to perform temporal segmentation on expert demonstrations $\mathcal{D}$. Specifically, given a task description and an expert video $\tau$, the VLM is prompted with the task description and uniformly sampled video frames (with original indices), and outputs {JSON-only} segments in the form \texttt{[\{stage\_number, start\_frame, end\_frame\}]}. We enforce monotonic, full-coverage boundaries and discard/re-query any ill-formed outputs. The VLM then identifies salient transition points and partitions $\tau$ into $K$ functional segments, where the start and end points of each segment correspond to specific frame indices in the original video. To ensure task logic consistency, we set a task-specific stage number $K$ using a robust aggregation over multiple reference demonstrations (e.g., mode/trimmed-mean over VLM-predicted segment counts) to reduce sensitivity to occasional over-/under-segmentation. We then assign each frame $o_t$ a categorical pseudo-label $y_t \in \{0, 1, \dots, K-1\}$ according to its segment, transforming raw visual observations into a structured sequence of sub-goals. We denote the resulting pseudo-labeled dataset as $ \mathcal{D}_{\text{label}}=\{(\tau,\mathbf{y})\}$ where $\mathbf{y}=\{y_1,\dots,y_T\}. $
This automated labeling provides scalable supervision for training the downstream temporal discriminator without costly manual frame-level annotation. Importantly, VLM-based segmentation is performed only during offline pre-training to generate pseudo-labels. We treat VLM-based segmentation as a weak supervision source rather than an oracle. STDR does not require precise boundary alignment: the downstream stage discriminator is trained to capture the coarse, order-consistent task logic under noisy pseudo labels, and online inference relies solely on the learned reward model (with OOD-aware gating and grasp-transition verification) rather than VLM outputs.

\subsection{Reward Model Architecture}

\paragraph{Stage Discriminator}


To accurately capture the logical progression of multi-stage tasks, we train a Stage Discriminator. 
At time step $t$, the discriminator takes as input a sliding window of observations with length $L$,
\[
\mathcal{O}_t = \{o_{t-L+1}, \dots, o_t\}.
\]
Unlike standard feed-forward architectures that process frames independently, we employ an LSTM~\cite{hochreiter1997long} to encode $\mathcal{O}_t$ and extract temporal dependencies that are essential in manipulation tasks. This sequence-based design enables the model to detect key ``transition points'' in the task flow, such as the transition from approaching the target to initiating the grasp. During training, the discriminator is supervised using the pseudo-labeled dataset $\mathcal{D}_{\text{label}}$, learning to map each temporal window to a corresponding stage index $k_t \in \{0,1,\dots,K-1\}$ and thereby capturing the underlying temporal logic in expert demonstrations.

Once pre-trained, the discriminator serves as an autonomous stage estimator during online reinforcement learning, providing real-time stage inference. To improve robustness under distributional shifts, we further integrate an auxiliary Variational Autoencoder (VAE)~\cite{kingma2022autoencodingvariationalbayes} for OOD detection. To contrast the expert manifold against task-irrelevant behaviors, we augment training with five short random-policy rollouts per task as negative examples. At inference time, the VAE monitors the reconstruction error of $\mathcal{O}_t$; if it exceeds a threshold $\delta$, we conservatively reset the stage prediction to the initial stage ($k_t \leftarrow 0$), suppressing spurious reward escalation under OOD observations.

Notably, in many manipulation tasks, the gripper occupies only a small portion of the visual observation, which makes it difficult for global encoders to reliably recognize subtle interaction states. To address this limitation, we introduce a multi-layer perceptron (MLP)-based Grasping Regulation module. This module acts as a visual verification gate for grasp-related transitions: the stage index $k_t$ is allowed to advance from the grasp stage to subsequent stages (e.g., moving or placing) only when the module predicts a stable grasp from visual features. The module is enabled only for tasks that require gripper-based grasping.

Finally, we define the stage reward as 
\begin{equation}
    R_{\text{stage}} = \frac{k_t}{K},
\end{equation}
which maps the inferred stage to $K$ discrete progress points $\{0, 1/K, \dots, (K-1)/K\}$. Overall, this multi-component design enforces logically consistent rewards by grounding stage transitions in temporally coherent representations, OOD-aware safeguards, and verified grasping cues.

\paragraph{Intra-stage Progress Estimator}

We introduce an intra-stage progress reward $R_{\text{prog}}$ to provide dense guidance within each sub-stage. Specifically, the reward estimates a normalized progress value $p_{\text{prog}} \in [0,1]$ within the current stage. Training data are derived from expert demonstrations segmented by a VLM and tagged with stage labels; we train a single progress estimator shared across stages while conditioning it on the current stage embedding.

{Progress network.} We build a deep visual progress network $\Phi$, consisting of a ResNet~\cite{he2016deep} backbone and a fusion MLP, to predict $p_t = \Phi(o_t, e_k)$, where $e_k$ denotes the embedding of the current stage index. To prevent $\Phi$ from overfitting to success-only data and erroneously rewarding failure states, we apply the Video Rewind technique from the ReWiND study \cite{zhang2025rewind}. Since the relevant visual cues vary across stages, we adopt a Feature-wise Linear Modulation (FiLM) ~\cite{perez2018film} modulation mechanism to dynamically condition the visual backbone on $e_k$. Concretely, intermediate visual features are transformed via stage-conditioned affine parameters, allowing a single vision stream to adaptively focus on stage-relevant information. 

{Training objective.} We train $\Phi$ with a composite loss over expert trajectories $\{o_t\}_{t=0}^T$. In addition to a regression loss $\mathcal{L}_{reg}$ (defined below), we employ three complementary constraints:
(1) \emph{Pairwise ranking loss} enforces temporal directionality by sampling $i < j$ from the same trajectory and applying a margin $m$:
\begin{equation}
\mathcal{L}_{rank} = \mathbb{E}_{i, j \sim \tau, i < j} \left[ \max(0, m - (p_j - p_i)) \right].
\end{equation}
(2) \emph{Isotonic regularization} penalizes any decrease in progress between consecutive steps to encourage monotonicity:
\begin{equation}
\mathcal{L}_{iso} = \mathbb{E}_{t \in [0, T-1]} \left[ \max(0, p_t - p_{t+1}) \right].
\end{equation}
(3) \emph{Visual consistency loss} uses dual-stream augmentation to enforce invariance to visual perturbations, by comparing strong ($o_t^s$) and weak ($o_t^w$) augmented versions of the same observation $o_t$:
\begin{equation}
\mathcal{L}_{cons} = \mathbb{E}_{o_t \sim \tau} \left[ \left\| \Phi(o_t^s, e_k) - \Phi(o_t^w, e_k) \right\|_1 \right].
\end{equation}
For regression supervision, we define an intra-stage local temporal target $\tilde{p}_t = \frac{t}{T-1} \in [0,1]$, where $t$ denotes the relative index within the current sub-stage of total length $T$. The regression loss $\mathcal{L}_{reg}$ is then defined as:
\begin{equation}
\mathcal{L}_{reg} = \mathbb{E}_{t \sim \tau} \left[ \left\| \Phi(o_t, e_k) - \tilde{p}_t \right\|_2^2 \right].
\end{equation}
The final pre-training objective is a weighted sum:
\begin{equation}
\mathcal{L}_{total} = \lambda_{reg}\mathcal{L}_{reg} + \lambda_{rank}\mathcal{L}_{rank} + \lambda_{iso}\mathcal{L}_{iso} + \lambda_{cons}\mathcal{L}_{cons},
\end{equation}
where the weighting coefficients $\lambda$ are hyperparameters used to balance the diverse loss components.

{OOD-aware fusion.} To mitigate reward hacking in unfamiliar states, we incorporate a Mahalanobis-distance~\cite{mahalanobis2018generalized}-based OOD detector to adaptively balance the regressed progress $p_{reg}$ and the conservative retrieval-based estimate $p_{near}$. First, we maintain a stage-specific retrieval database that stores expert ``window-averaged features from fused feature layer $h$ -- temporal progress'' pairs. Let $d_{\text{ood}}$ denote the Mahalanobis distance of the current observation on feature layer $h$ to the expert manifold. We compute an adaptive mixing weight $g(d_\text{ood}) = \sigma(-\kappa d_{\text{ood}})$, serving as a monotonically decreasing gate. When the state is close to the expert distribution, $g(d_{\text{ood}})$ is large, and the model relies more on the regressed progress $p_{reg}$; when it is far from the expert distribution, the model falls back to the conservative retrieval-based progress $p_{near}$ obtained via nearest-neighbor alignment. The fused progress is first calculated as a weighted estimate $\hat{p}$:
\begin{equation}
    \hat{p}_{\text{prog}} = g(d_{\text{ood}}) \cdot p_{\text{reg}} + (1 - g(d_{\text{ood}})) \cdot p_{\text{near}}
\end{equation}
Then, the final fused estimate $p_{\text{prog}}$ is defined as:
\begin{equation}
    p_{\text{prog}} = 
    \begin{cases} 
    0 & \text{if } d_{\text{ood}} > \eta, \\
    \hat{p}_{\text{prog}} & \text{otherwise},
    \end{cases}
\end{equation}
where $\eta$ is a predefined distance threshold used to enforce a progress upper bound when the state significantly deviates from the expert distribution. Finally, the intra-stage reward is defined as 
\begin{equation}
    R_{\text{prog}} = \frac{p_{\text{prog}} }{K}.
\end{equation}
This scaling logic confines the dense signal within adjacent stage points, ensuring that the global incentive landscape is always dominated by discrete stage transitions.

{Overall reward.} Finally, the synthesized reward is
\begin{equation}
R_\text{STDR} =
\begin{cases}
1.0 & \text{if } \text{success}, \\
R_\text{stage} + R_\text{prog} & \text{otherwise}.
\end{cases}
\end{equation}
This formulation yields a globally monotonic incentive landscape over the task horizon by combining discrete stage transitions with continuous intra-stage progress.

\subsection{System Summary}

\paragraph{Reward Inference} To compute the reward at timestep $t$, our framework adopts a dual reward-model architecture consisting of a {Stage Discriminator} and an {Intra-stage Progress Estimator}. The inference pipeline proceeds as follows: (1) an observation sequence $\mathcal{O}_t$ of length $L$ is first processed by the LSTM-based stage discriminator to infer the current high-level stage index $k_t \in \{0, \dots, K-1\}$; (2) this stage index is transformed into a stage embedding to dynamically modulate the visual backbone of the progress estimator via a FiLM mechanism, allowing the model to focus on stage-specific visual cues; (3) the estimator outputs a regressed progress value $p_\text{reg} \in [0, 1]$, which is then fused with a retrieval-based alignment $p_\text{near}$ using a Mahalanobis distance-based OOD detector to produce the final progress estimate $p_\text{prog}$; (4) finally, the system calculates stage progress reward $R_\text{stage}$ based on the stage index $k_t$ and intra-stage progress reward $R_\text{prog}$ (in non-success steps) to synthesize the final dense reward signal $R_\text{STDR}$.
In our experiments, we used unified hyperparameters for all tasks, including the Mahalanobis distance threshold $\eta$, the gating parameter $\kappa$, and the VAE reconstruction error threshold $\delta$, to demonstrate the robustness of the STDR framework.

\paragraph{Training} Our framework is implemented in two primary phases: {STDR pre-training} and {reward-driven online learning}. The objective of the first phase is to learn a structured reward function by distilling task-specific temporal knowledge from unstructured expert videos. In the second phase, this pre-trained model is utilized to train a RL agent from scratch through online interaction. Central to this process is the STDR, which maps visual progress into a dense, continuous feedback signal. 

\section{EXPERIMENTS}

In this section, we evaluate the effectiveness of the proposed STDR framework through a series of experiments. Our study aims to address three core questions: 
\begin{enumerate}
    \item Can STDR significantly improve sample efficiency and final success rates in challenging long-horizon manipulation tasks compared to existing reward shaping methods and sparse reward baselines? 
    \item To what extent can the learned reward model generalize to real-world visual observations, and can it provide more reliable guidance signals than representation-based methods like VIP and LIV in the presence of environmental noise and the sim-to-real domain gap?
    \item What are the specific contributions of the core components within the STDR architecture---namely the Stage Discriminator, the Intra-stage Progress Estimator, and the Grasping Regulation module---to the overall performance?
\end{enumerate}

Specifically, we first introduce the experimental setups across three simulation platforms and analyze the quantitative performance of STDR against various baselines in Sec.~\ref{sec:simulation}. We then investigate the practical feasibility of our framework and the quality of reward signals under real-world visual complexity in Sec.~\ref{sec:real_robot}. Finally, we conduct detailed ablation studies in Sec.~\ref{sec:ablation} to evaluate the necessity of each core module within the STDR architecture. 

\subsection{Simulation Experiments}\label{sec:simulation}

\begin{figure*}[t] 
    \centering
    \includegraphics[width=1\textwidth]{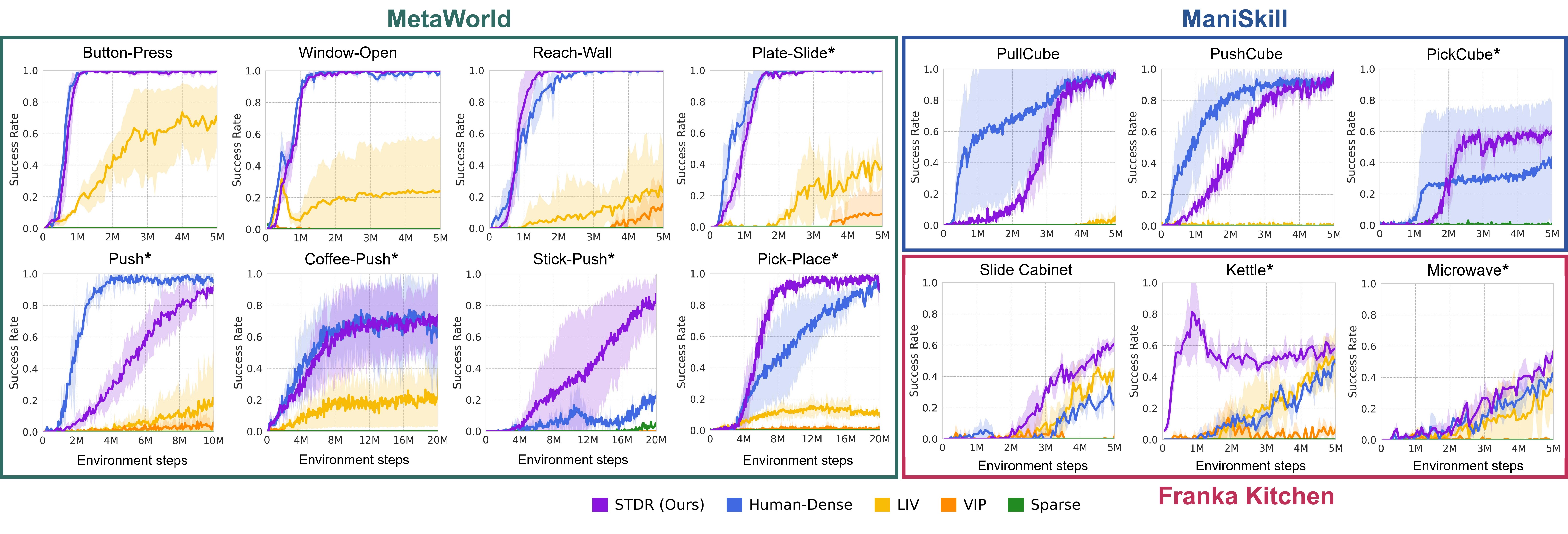}
    \caption{Training success rates across 14 robotic manipulation tasks. Tasks are grouped by environment: MetaWorld (first 8 plots), ManiSkill (top-right 3 plots), and Franka Kitchen (bottom-right 3 plots). Solid lines represent the mean success rate across 4 independent runs(different seeds), and shaded areas denote the standard deviation. Note: Task titles marked with an asterisk ($*$) incorporate the Grasping Regulation module.}
    \label{fig:main_results}
\end{figure*}

\paragraph{Experimental Setup}
To comprehensively evaluate the performance of the STDR framework, we benchmark our method across 14 tasks selected from three distinct simulation platforms. We utilize MetaWorld \cite{mclean2025metaworld} (8 tasks) to assess the generalization and spatial adaptability of the reward model under randomized initializations. For tasks requiring high precision, ManiSkill \cite{gu2023maniskill2} (3 tasks) provides high-fidelity physics to verify fine-grained guidance and spatial invariance. Finally, the Franka Kitchen \cite{gymnasium_robotics2023github} (3 tasks) environment, characterized by cluttered backgrounds and diverse joint constraints, serves to evaluate the feature selection capabilities of the temporal stage discriminator under high visual complexity.

For reward model training, 30 expert demonstrations are prepared per task. Data for MetaWorld and ManiSkill are rendered via official expert scripts, while Franka Kitchen trajectories are extracted from the D4RL \cite{fu2020d4rl} dataset. All visual observations are encoded using a pre-trained ResNet \cite{he2016deep} backbone. Policy optimization is consistently performed using Proximal Policy Optimization (PPO) \cite{schulman2017proximal}. Based on framework compatibility, we employ Stable Baselines3 (SB3) \cite{stable-baselines3} for MetaWorld and Franka Kitchen, and native official implementation for ManiSkill to ensure baseline performance aligns with official standards.

\paragraph{Baselines}
We compare STDR against four representative reward mechanisms:
\begin{enumerate}
    \item Sparse (Official): The raw binary success reward provided by the environment. This comparison aims to verify the performance gain of adding our pre-trained reward model in the absence of any intermediate guidance.
    \item Human-Dense: A manually engineered dense reward function based on the ground-truth state. We utilize the official reward functions provided by the Meta-World \cite{mclean2025metaworld} and ManiSkill \cite{gu2023maniskill2} environments, whereas for Franka Kitchen\cite{gymnasium_robotics2023github} , we developed custom reward functions as no official implementations are available. This comparison aims to verify whether our visual reward model can reach or even exceed the performance upper bound of hand-tuned rewards without requiring expert domain knowledge for tuning.
    \item VIP \cite{ma2022vip}: This method utilizes smooth visual representations learned from large-scale human videos, defining the reward as the negative distance between the current state and the goal state in the embedding space. This comparison aims to verify the advantages of introducing explicit temporal logic sequences and phased goals over implicit preference learning that relies solely on single goal image guidance in long-horizon tasks.
    \item LIV \cite{ma2023liv}: This method generates rewards based on the semantic similarity between current observations and target images via cross-modal alignment. This comparison aims to benchmark STDR against it and verify the superiority of our method in visual representation learning tasks.
\end{enumerate}

\paragraph{Result}

The training results across 14 tasks are illustrated in Fig.~\ref{fig:main_results}. We first analyze the performance across the three simulation platforms:

\begin{itemize}
    \item \textit{MetaWorld Analysis:} STDR exhibits rapid convergence across the 8 tasks in MetaWorld, typically reaching a 1.0 success rate within 1M to 2M steps for simpler tasks. Notably, in several high-difficulty tasks such as \textit{Pick-Place}, \textit{Stick-Push}, \textit{etc.}, STDR matches or even surpasses the performance of \textit{Human-Dense} rewards.
    \item \textit{ManiSkill Analysis:} In the ManiSkill environment requiring high-precision control, baseline methods such as \textit{LIV} and \textit{VIP} show negligible progress. In contrast, STDR maintains stable and superior performance through its temporal stage decomposition mechanism, validating its reliability in fine-grained robotic manipulation.
    \item \textit{Franka Kitchen Analysis:} Results prove that STDR remains effective in complex scenes with multiple co-existing objects, successfully identifying key task transitions and outperforming all baseline methods.
\end{itemize}

In summary, under identical visual-only observation settings, our method outperforms both \textit{LIV} and \textit{VIP} baselines across all 14 tasks. Compared to methods relying on global semantic similarity, the stage-transition logic in STDR provides clearer gradients for spatial alignment and superior representation capabilities for long-horizon tasks. Furthermore, despite relying solely on visual input, STDR surpasses state-based \textit{Human-Dense} rewards in several high-difficulty tasks, highlighting its superior guidance efficacy over manually tuned functions without requiring domain expertise.

\paragraph{Data Efficiency and Performance Analysis}
To further validate the data efficiency of STDR, we conducted a specialized evaluation in the ManiSkill environment. We compared the policy performance of RL guided by different reward signals against the pure offline imitation learning method, Diffusion Policy (DP)\cite{chi2024diffusionpolicy}, by testing the best checkpoints over 50 evaluation episodes. Specifically, to ensure a rigorous and fair comparison, the DP baseline was trained using the exact same set of expert demonstrations as those employed for our reward model. The results in Table~\ref{tab:maniskill_data_efficiency} show that STDR achieves higher data utilization efficiency. In these high-precision tasks, using the same demonstrations, STDR achieves better generalization compared to Diffusion Policy. Furthermore, the performance of STDR far exceeds visual representation baselines such as LIV \cite{ma2023liv} and VIP \cite{ma2022vip}.


\begin{table}[h]
\centering
\caption{Evaluation of performance and data efficiency in ManiSkill. Results represent the mean and standard deviation of success rates over 50 evaluation episodes using the best-performing checkpoints.}
\label{tab:maniskill_data_efficiency}
\resizebox{\columnwidth}{!}{%
\begin{tabular}{lccc}
\hline
\textbf{Method} & \textbf{PullCube} & \textbf{PushCube} & \textbf{PickCube} \\ \hline
STDR (Ours) & $\mathbf{0.959 \pm 0.029}$ & $\mathbf{0.943 \pm 0.031}$ & $0.632 \pm 0.029$ \\
DP & $0.120 \pm 0.038$ & $0.129 \pm 0.023$ & $0.060 \pm 0.035$ \\
LIV & $0.051 \pm 0.025$ & $0.034 \pm 0.011$ & $0.000 \pm 0.000$ \\
VIP & $0.000 \pm 0.000$ & $0.000 \pm 0.000$ & $0.000 \pm 0.000$ \\
Human-Dense & $0.955 \pm 0.021$ & $0.931 \pm 0.031$ & $\mathbf{0.961 \pm 0.044}$ \\
Sparse & $0.000 \pm 0.000$ & $0.000 \pm 0.000$ & $0.010 \pm 0.005$ \\ \hline
\end{tabular}%
}
\end{table}

\subsection{Real-World Evaluation} \label{sec:real_robot}

\begin{figure*}[t] 
    \centering
    \includegraphics[width=1\textwidth]{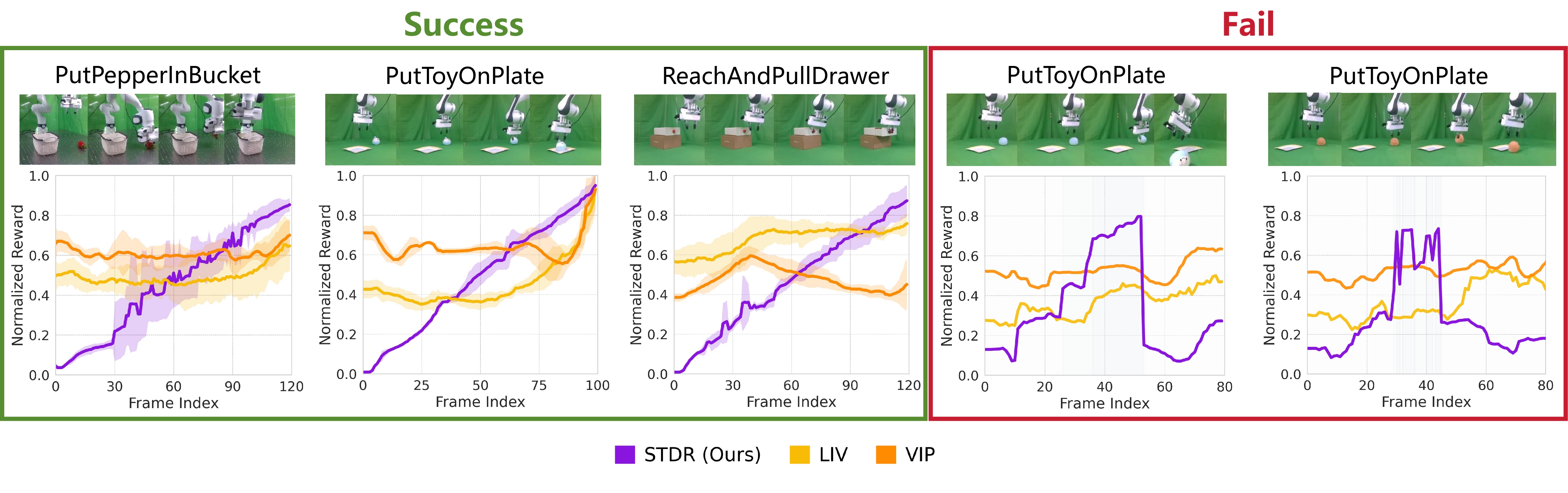} 
    \caption{Reward assignment of different methods during physical robotic manipulation tasks. The left panels illustrate the reward distribution on successful task videos, where solid lines denote the mean reward across 5 evaluation videos and the shaded areas indicate the standard deviation. The right panels display the reward assignment on failure videos, representing evaluations on individual video sequences. The x-axis represents the frame index, and the y-axis represents the normalized reward.} 
    \label{fig:real_world}
\end{figure*}

To verify the practical feasibility and cross-environment transferability of STDR, we conducted experiments on a physical robotic platform comprising a Franka arm and an Intel RealSense camera. For each task, we collected 15 successful real-world observation videos: 10 segments were used to train the stage-transition logic, and the remaining 5 segments were used for performance evaluation on unseen trajectories. Additionally, several failure videos were collected for robustness testing.

 For a rigorous comparison, we fine-tuned the VIP and LIV baselines via temporal contrastive learning (TCL) \cite{sermanet2018time} using the identical 10 expert demonstrations employed for STDR, while keeping their visual encoders frozen. The quantitative comparison of reward curves is shown in Fig.~\ref{fig:real_world}.

Evaluation on Successful Videos (Left) results show that the rewards assigned by STDR exhibit significant stability and an increasing trend, which is highly coupled with task progress. In contrast, although LIV \cite{ma2023liv} and VIP \cite{ma2022vip} maintain high rewards at the end of the task, they exhibit obvious signal plateaus in the intermediate stages and lack the discriminative gradients required to guide action sequences.

Evaluation on Failure Videos (Right) results indicate that baseline methods fail to detect task failure scenarios (e.g., an object dropping), still continuing to provide erroneous positive feedback. Conversely, STDR provides low reward signals that align with physical execution logic by accurately identifying the failure state.

This advantage stems from STDR's explicit temporal stage modeling, which ensures the reward signal is closely aligned with the physical execution logic, effectively overcoming background interference and sensor noise in real-world environments. These findings demonstrate that STDR provides a more accurate and reasonable guidance signal, showcasing its potential for real-world deployment.

\subsection{Ablation Study} \label{sec:ablation}

To further investigate the contributions of core components and the sensitivity to data scale within the STDR framework, we conduct a series of ablation experiments.
\subsubsection{Component-wise Ablation Analysis}

\begin{figure}[t] 
    \centering
    \includegraphics[width=0.46\textwidth]{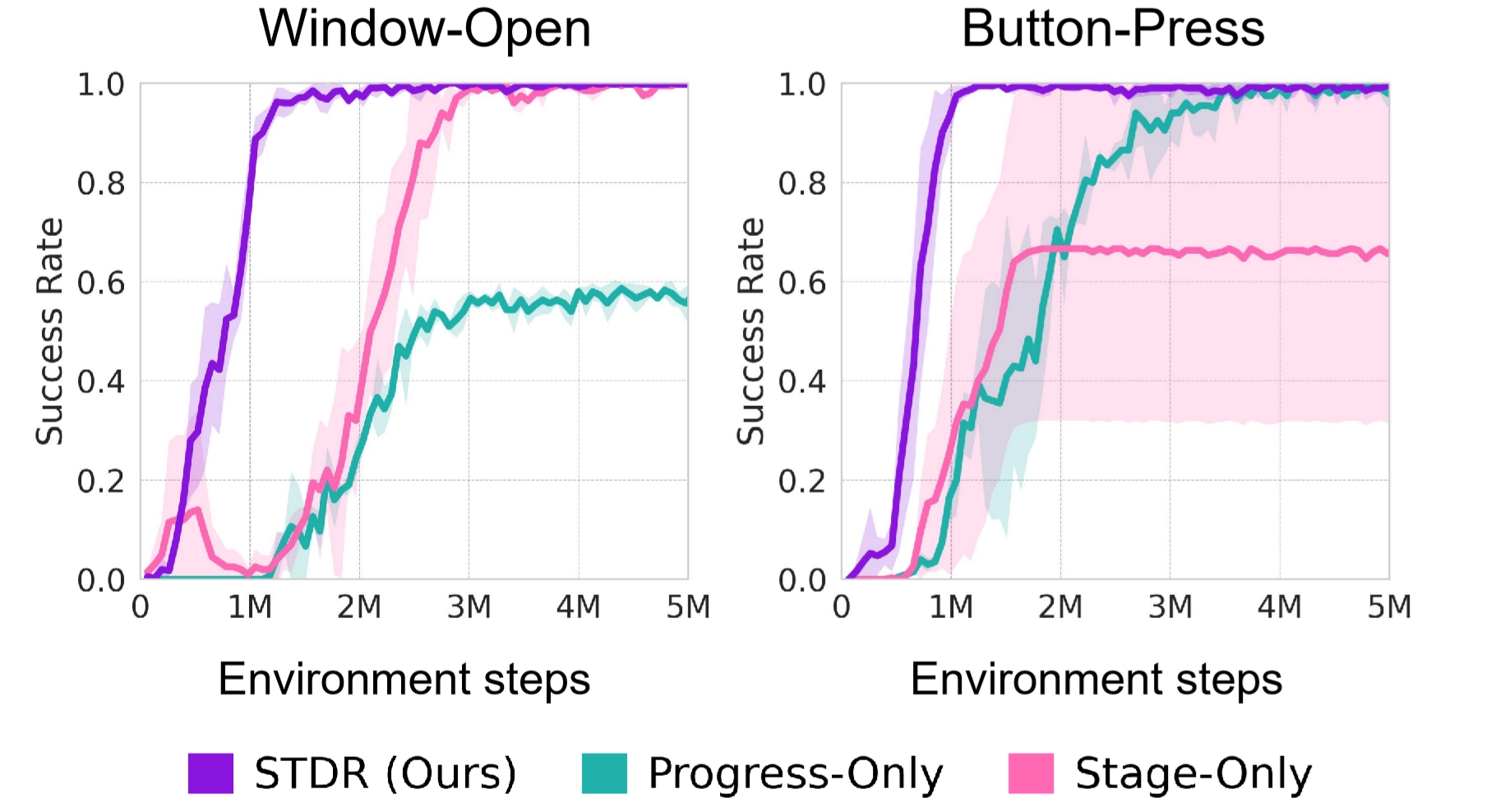} 
    \caption{Learning curves of component-wise ablation results on Window-Open (left) and Button-Press (right) tasks, measured by success rate. Solid lines and shaded regions represent the mean and standard deviation across four independent runs, respectively.} 
    \label{fig:Ablation1}
\end{figure}

We evaluate the full STDR model against two variants to verify the hierarchical reward structure:
\begin{itemize}
    \item Stage-Only (w/o Progress Estimator): This variant removes the intra-stage progress estimator, relying solely on discrete stage transitions for reward signals. This comparison evaluates the necessity of continuous guidance within sub-tasks for efficient exploration.
    \item Progress-Only (w/o Stage Discriminator): This variant eliminates explicit stage partitioning and relies on global progress estimation. This comparison highlights the importance of logical decomposition in solving long-horizon tasks.
\end{itemize}

The quantitative results of the component ablation are illustrated in Fig.~\ref{fig:Ablation1}. The comparison highlights the synergistic effect of the hierarchical reward components.

As shown in the training curves, the Stage-Only variant exhibits a significant learning delay and notably high variance. While it eventually reaches a high success rate in the Window-Open task, the progress is markedly slower. This suggests that without intra-stage gradients, the agent's exploration within each functional segment becomes inconsistent, leading to unstable performance.

Conversely, the Progress-Only variant shows a noticeable decrease in the final success rate. In the absence of an explicit logical stage structure, the agent struggles to maintain the correct execution sequence required for long-horizon manipulation. These findings validate that the combination of logical transitions and continuous guidance in STDR ensures a monotonic incentive landscape, which is essential for stable and efficient policy optimization.

\subsubsection{Impact of Grasping Regulation} 

\begin{figure}[t] 
    \centering
    \includegraphics[width=0.46\textwidth]{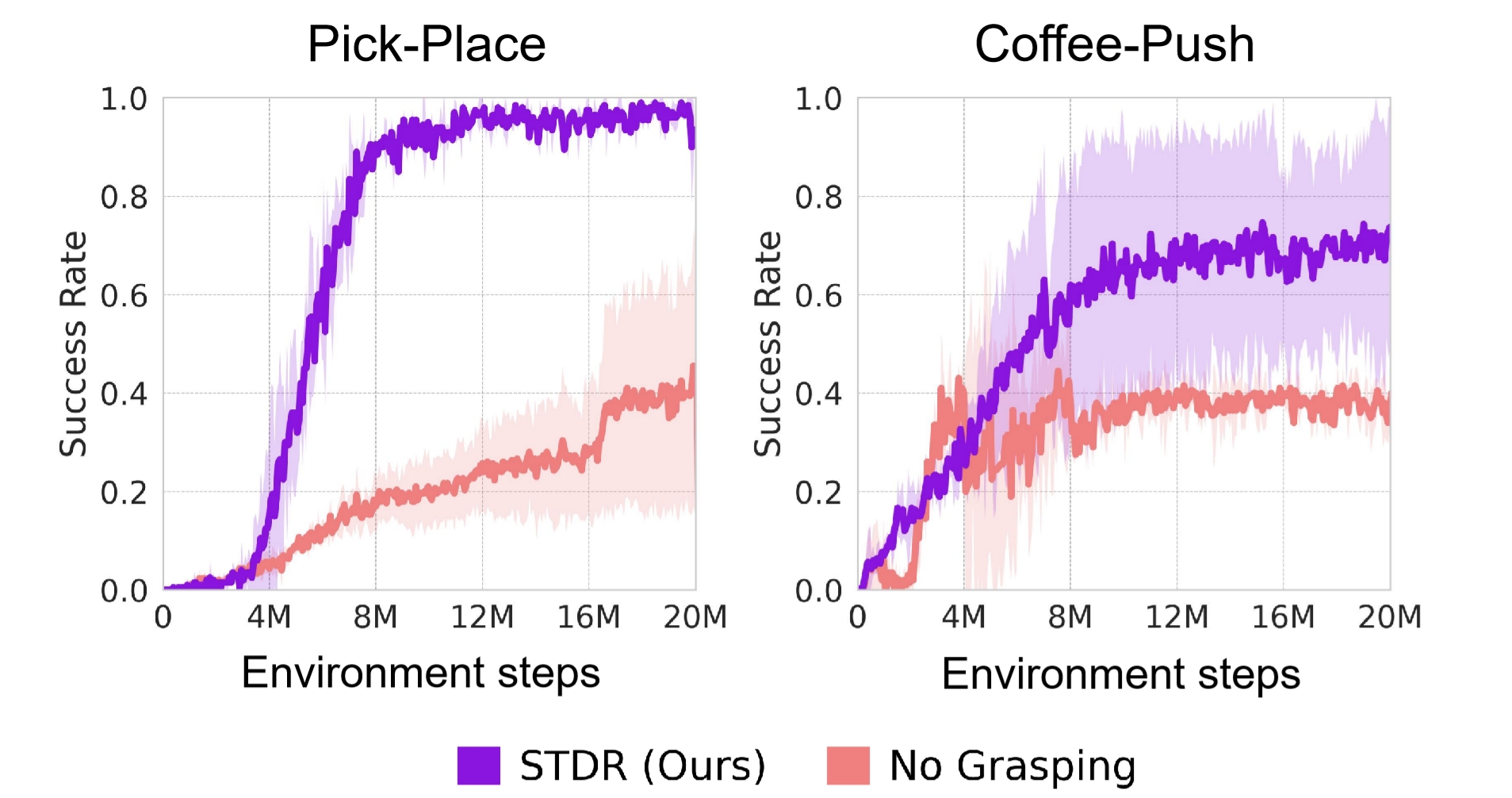} 
    \caption{Learning curves of ablation results on both the Pick-Place (left) and Coffee-Push (right) tasks. Solid lines and shaded regions represent the mean and standard deviation across four independent runs, respectively.} 
    \label{fig:Ablation2}
\end{figure}

For interaction tasks, we evaluate the role of the MLP-based grasping discriminator in regulating stage transitions:
\begin{itemize}
    \item No Grasping (w/o Grasping Regulation): In both  the Pick-Place and Coffee-Push tasks, we remove the verification gate so that stage transitions occur without physical stability checks, relying purely on spatial proximity.
\end{itemize}

The quantitative results are illustrated in Fig.~\ref{fig:Ablation2}. Our analysis reveals that removing this module leads to ``stage leakage,'' where the reward signal prematurely transitions to the next functional stage before the agent has securely grasped the object. Consequently, the agent often moves toward the target with an empty gripper or drops the object mid-air. These findings validate that incorporating a physical state discriminator is essential for aligning the incentive landscape with the actual task logic in complex manipulation environments.


\section{Conclusions and Limitations}
In this paper, we presented STDR, a visual reward learning framework designed to bridge the gap between unstructured video demonstrations and structured dense reward signals for long-horizon robotic manipulation. By distilling functional stage logic from expert videos and processing real-time continuous observation streams during reinforcement learning, STDR provides a scalable solution to the traditional reward engineering bottleneck. Our comprehensive evaluation across 14 tasks in three diverse simulation environments demonstrates that STDR not only consistently outperforms existing task-agnostic baselines but also surpasses hand-crafted rewards in several challenging scenarios. 

Furthermore, the successful validation of STDR on real-world visual data confirms the robustness and generalization potential of our vision-driven components, specifically the grasping regulation module which serves as a reliable verification gate for stage transitions. Overall, STDR offers a promising path toward autonomous robotic learning with minimal human intervention. Despite these achievements, STDR has some limitations that merit further investigation. Currently, the reward model relies solely on visual observations. For tasks requiring richer perception, incorporating additional signals—such as proprioceptive state information and tactile feedback—could be a necessary and promising direction. Moreover, STDR is most suitable for tasks with basically consistent stage sequences, but it is less adaptable to tasks with exchangeable sequences. Future work will therefore focus on extending STDR to integrate multi-modal sensory inputs and to enhance its robustness and adaptability. Beyond improving the reward model itself, another important direction is to investigate training full reinforcement learning policies directly on physical robots, building upon the stable reward signals demonstrated in our real-world experiments.


\bibliographystyle{IEEEtran}
\bibliography{ref}

@article{bellman1957markovian,
  title={A Markovian decision process},
  author={Bellman, Richard},
  journal={Journal of mathematics and mechanics},
  pages={679--684},
  year={1957},
  publisher={JSTOR}
}

@article{qwen3,
    title={Qwen3 Technical Report}, 
    author={An Yang and Anfeng Li and Baosong Yang and others},
    journal = {arXiv preprint arXiv:2505.09388},
    year={2025}
}

@article{hochreiter1997long,
  title={Long short-term memory},
  author={Hochreiter, Sepp and Schmidhuber, J{\"u}rgen},
  journal={Neural computation},
  volume={9},
  number={8},
  pages={1735--1780},
  year={1997},
  publisher={MIT press}
}

@article{ma2023liv,
  title   = {{{LIV}}: Language-Image Representations and Rewards for Robotic Control},
  author  = {Ma, Yecheng Jason and Liang, William and Som, Vaidehi and Kumar, Vikash and Zhang, Amy and Bastani, Osbert and Jayaraman, Dinesh},
  journal = {arXiv preprint arXiv:2306.00958},
  year    = {2023}
}

@article{ma2022vip,
  title   = {{{VIP}}: Towards Universal Visual Reward and Representation via Value-Implicit Pre-Training},
  author  = {Ma, Yecheng Jason and Sodhani, Shagun and Jayaraman, Dinesh and Bastani, Osbert and Kumar, Vikash and Zhang, Amy},
  journal = {arXiv preprint arXiv:2210.00030},
  year    = {2022}
}

@InProceedings{perez2018film,
  title={FiLM: Visual Reasoning with a General Conditioning Layer},
  author={Ethan Perez and Florian Strub and Harm de Vries and Vincent Dumoulin and Aaron C. Courville},
  booktitle={AAAI},
  year={2018}
}

@misc{fu2020d4rl,
    title={D4RL: Datasets for Deep Data-Driven Reinforcement Learning},
    author={Justin Fu and Aviral Kumar and Ofir Nachum and George Tucker and Sergey Levine},
    year={2020},
    eprint={2004.07219},
    archivePrefix={arXiv},
    primaryClass={cs.LG}
}

@article{schulman2017proximal,
  title={Proximal policy optimization algorithms},
  author={Schulman, John and Wolski, Filip and Dhariwal, Prafulla and Radford, Alec and Klimov, Oleg},
  journal={arXiv preprint arXiv:1707.06347},
  year={2017}
}

@article{stable-baselines3,
  author  = {Antonin Raffin and Ashley Hill and Adam Gleave and Anssi Kanervisto and Maximilian Ernestus and Noah Dormann},
  title   = {Stable-Baselines3: Reliable Reinforcement Learning Implementations},
  journal = {Journal of Machine Learning Research},
  year    = {2021},
  volume  = {22},
  number  = {268},
  pages   = {1-8},
  url     = {http://jmlr.org/papers/v22/20-1364.html}
}

@inproceedings{gu2023maniskill2,
  title={ManiSkill2: A Unified Benchmark for Generalizable Manipulation Skills},
  author={Gu, Jiayuan and Xiang, Fanbo and Li, Xuanlin and Ling, Zhan and Liu, Xiqiang and Mu, Tongzhou and Tang, Yihe and Tao, Stone and Wei, Xinyue and Yao, Yunchao and Yuan, Xiaodi and Xie, Pengwei and Huang, Zhiao and Chen, Rui and Su, Hao},
  booktitle={International Conference on Learning Representations},
  year={2023}
}

@inproceedings{
mclean2025metaworld,
title={Meta-World+: An Improved, Standardized, {RL} Benchmark},
author={Reginald McLean and Evangelos Chatzaroulas and Luc McCutcheon and Frank R{\"o}der and Tianhe Yu and Zhanpeng He and K.R. Zentner and Ryan Julian and J K Terry and Isaac Woungang and Nariman Farsad and Pablo Samuel Castro},
booktitle={The Thirty-ninth Annual Conference on Neural Information Processing Systems Datasets and Benchmarks Track},
year={2025},
url={https://openreview.net/forum?id=1de3azE606}
}

@misc{gymnasium_robotics2023github,
  author = {Rodrigo de Lazcano and Kallinteris Andreas and Jun Jet Tai and Seungjae Ryan Lee and Jordan Terry},
  title = {Gymnasium Robotics},
  url = {http://github.com/Farama-Foundation/Gymnasium-Robotics},
  version = {1.4.0},
  year = {2024},
}

@misc{kingma2022autoencodingvariationalbayes,
      title={Auto-Encoding Variational Bayes}, 
      author={Diederik P Kingma and Max Welling},
      year={2022},
      eprint={1312.6114},
      archivePrefix={arXiv},
      primaryClass={stat.ML},
      url={https://arxiv.org/abs/1312.6114}, 
}

@inproceedings{he2016deep,
  author={He, Kaiming and Zhang, Xiangyu and Ren, Shaoqing and Sun, Jian},
  booktitle={Proceedings of the IEEE Conference on Computer Vision and Pattern Recognition (CVPR)}, 
  title={Deep Residual Learning for Image Recognition}, 
  year={2016},
  pages={770-778},
  doi={10.1109/CVPR.2016.90}
}

@article{mahalanobis2018generalized,
  title={On the generalized distance in statistics},
  author={Mahalanobis, Prasanta Chandra},
  journal={Sankhy{\=a}: The Indian Journal of Statistics, Series A (2008-)},
  volume={80},
  pages={S1--S7},
  year={2018},
  publisher={JSTOR}
}

@misc{luo2024hilserl,
      title={Precise and Dexterous Robotic Manipulation via Human-in-the-Loop Reinforcement Learning},
      author={Jianlan Luo and Charles Xu and Jeffrey Wu and Sergey Levine},
      year={2024},
      eprint={2410.21845},
      archivePrefix={arXiv},
      primaryClass={cs.RO}
}

@article{levine2016end,
  title={End-to-end training of deep visuomotor policies},
  author={Levine, Sergey and Finn, Chelsea and Darrell, Trevor and Abbeel, Pieter},
  journal={Journal of Machine Learning Research},
  volume={17},
  number={39},
  pages={1--40},
  year={2016}
}

@inproceedings{nair2018overcoming,
  title={Overcoming exploration in reinforcement learning with demonstrations},
  author={Nair, Ashvin and McGrew, Bob and Andrychowicz, Marcin and Zaremba, Wojciech and Abbeel, Pieter},
  booktitle={2018 IEEE international conference on robotics and automation (ICRA)},
  pages={6292--6299},
  year={2018},
  organization={IEEE}
}

@inproceedings{riedmiller2018learning,
  title={Learning by playing solving sparse reward tasks from scratch},
  author={Riedmiller, Martin and Hafner, Roland and Lampe, Thomas and Neunert, Michael and Degrave, Jonas and Wiele, Tom and Mnih, Vlad and Heess, Nicolas and Springenberg, Jost Tobias},
  booktitle={International conference on machine learning},
  pages={4344--4353},
  year={2018},
  organization={PMLR}
}

@inproceedings{ng1999policy,
  title={Policy invariance under reward transformations: Theory and application to reward shaping},
  author={Ng, Andrew Y and Harada, Daishi and Russell, Stuart},
  booktitle={Icml},
  volume={99},
  pages={278--287},
  year={1999},
  organization={Citeseer}
}

@inproceedings{abbeel2004apprenticeship,
  title={Apprenticeship learning via inverse reinforcement learning},
  author={Abbeel, Pieter and Ng, Andrew Y},
  booktitle={Proceedings of the twenty-first international conference on Machine learning},
  pages={1},
  year={2004}
}

@article{amodei2016concrete,
  title={Concrete problems in AI safety},
  author={Amodei, Dario and Olah, Chris and Steinhardt, Jacob and Christiano, Paul and Schulman, John and Man{\'e}, Dan},
  journal={arXiv preprint arXiv:1606.06565},
  year={2016}
}

@inproceedings{brown2019extrapolating,
  title={Extrapolating beyond suboptimal demonstrations via inverse reinforcement learning from observations},
  author={Brown, Daniel and Goo, Wonjoon and Nagarajan, Prabhat and Niekum, Scott},
  booktitle={International conference on machine learning},
  pages={783--792},
  year={2019},
  organization={PMLR}
}

@article{rajeswaran2017learning,
  title={Learning complex dexterous manipulation with deep reinforcement learning and demonstrations},
  author={Rajeswaran, Aravind and Kumar, Vikash and Gupta, Abhishek and Vezzani, Giulia and Schulman, John and Todorov, Emanuel and Levine, Sergey},
  journal={arXiv preprint arXiv:1709.10087},
  year={2017}
}

@misc{escoriza2025multistage,
      title={Multi-Stage Manipulation with Demonstration-Augmented Reward, Policy, and World Model Learning},
      author={Adrià López Escoriza and Nicklas Hansen and Stone Tao and Tongzhou Mu and Hao Su},
      booktitle={International Conference on Machine Learning (ICML)},
      year={2025},
}

@article{liu2022inferring,
  title={Inferring human-robot performance objectives during locomotion using inverse reinforcement learning and inverse optimal control},
  author={Liu, Wentao and Zhong, Junmin and Wu, Ruofan and Fylstra, Bretta L and Si, Jennie and Huang, He Helen},
  journal={IEEE Robotics and Automation Letters},
  volume={7},
  number={2},
  pages={2549--2556},
  year={2022},
  publisher={IEEE}
}

@inproceedings{chen2023fast,
  title={Fast lifelong adaptive inverse reinforcement learning from demonstrations},
  author={Chen, Letian and Jayanthi, Sravan and Paleja, Rohan R and Martin, Daniel and Zakharov, Viacheslav and Gombolay, Matthew},
  booktitle={Conference on Robot Learning},
  pages={2083--2094},
  year={2023},
  organization={PMLR}
}

@article{chi2024diffusionpolicy,
	author = {Cheng Chi and Zhenjia Xu and Siyuan Feng and Eric Cousineau and Yilun Du and Benjamin Burchfiel and Russ Tedrake and Shuran Song},
	title ={Diffusion Policy: Visuomotor Policy Learning via Action Diffusion},
	journal = {The International Journal of Robotics Research},
	year = {2024},
}

@inproceedings{sermanet2018time,
  title={Time-contrastive networks: Self-supervised learning from video},
  author={Sermanet, Pierre and Lynch, Corey and Chebotar, Yevgen and Hsu, Jasmine and Jang, Eric and Schaal, Stefan and Levine, Sergey and Brain, Google},
  booktitle={2018 IEEE international conference on robotics and automation (ICRA)},
  pages={1134--1141},
  year={2018},
  organization={IEEE}
}

@inproceedings{
      zhang2025rewind,
      title={ReWi{ND}: Language-Guided Rewards Teach Robot Policies without New Demonstrations},
      author={Jiahui Zhang and Yusen Luo and Abrar Anwar and Sumedh Anand Sontakke and Joseph J Lim and Jesse Thomason and Erdem Biyik and Jesse Zhang},
      booktitle={9th Annual Conference on Robot Learning},
      year={2025},
      url={https://openreview.net/forum?id=XjjXLxfPou}
    }

\addtolength{\textheight}{-12cm}   

\end{document}